\documentclass[11pt]{article}

\usepackage[final]{acl}

\usepackage{times}
\usepackage{latexsym}

\usepackage[T1]{fontenc}
\usepackage[utf8]{inputenc}

\usepackage{microtype}

\usepackage{inconsolata}

\usepackage{graphicx}

\usepackage{microtype}
\usepackage{hyperref}
\usepackage{url}
\usepackage{booktabs}
\usepackage{graphicx}
\usepackage{subfigure}
\usepackage{multirow}

\usepackage{amsmath}
\usepackage{amssymb}
\usepackage{mathtools}
\usepackage{amsthm}

\usepackage[textsize=small]{todonotes}

\usepackage[most]{tcolorbox}
\usepackage{xcolor}
\usepackage{listings}
\usepackage{array}
\usepackage{tabularx}
\usepackage{algorithm}
\usepackage{algorithmic}
\usepackage[table]{xcolor}
\usepackage{colortbl, longtable}
\usepackage{wrapfig}
\usepackage{cleveref}
\usepackage{subcaption}
\usepackage{wrapfig}
\usepackage{enumitem}

\definecolor{Mycolor-red}{HTML}{FCEAEA} % light red

\definecolor{pinkframe}{HTML}{D26AA8}
\definecolor{pinktitle}{HTML}{F8D7EA}
\definecolor{pinkbg}{HTML}{FFF5FB}

\definecolor{myblueframe}{HTML}{4C8DFF} % 边框：浅蓝
\definecolor{mybluetitle}{HTML}{DCEAFF} % 标题栏：很浅蓝
\definecolor{mybluebg}{HTML}{F5F9FF}    % 内容底：极浅蓝

\definecolor{grayframe}{HTML}{7A7A7A}
\definecolor{graytitle}{HTML}{EDEDED}
\definecolor{graybg}{HTML}{F7F7F7}

\newtcblisting{promptbox}[2][]{
  enhanced,
  colframe=grayframe,
  colback=graybg,
  colbacktitle=graytitle,
  coltitle=black,
  fonttitle=\bfseries,
  title={#2},
  boxrule=0.8pt,
  arc=3pt,
  left=8pt,right=8pt,top=6pt,bottom=8pt,
  listing only,
  listing options={
    basicstyle=\ttfamily\small,
    breaklines=true,
    columns=fullflexible
  },
  #1
}

\newtcolorbox{promptboxmath}[1]{
  enhanced,
  colframe=grayframe,
  colback=graybg,
  colbacktitle=graytitle,
  coltitle=black,
  fonttitle=\bfseries,
  title={#1},
  boxrule=0.8pt,
  arc=3pt,
  left=8pt,right=8pt,top=6pt,bottom=8pt
}

\newtcblisting{steerbox}[1]{
  enhanced,
  colframe=myblueframe,
  colback=mybluebg,
  colbacktitle=mybluetitle,
  coltitle=black,
  fonttitle=\bfseries,
  title=#1,
  boxrule=0.8pt,
  arc=2pt,
  left=6pt,right=6pt,top=6pt,bottom=6pt,
  listing only,
  listing options={
    basicstyle=\ttfamily\small,
    breaklines=true,
    columns=fullflexible
  }
}

\newtcolorbox{skillcard}[1]{
  enhanced,
  colframe=pinkframe,
  colback=pinkbg,
  colbacktitle=pinktitle,
  coltitle=black,
  title=#1,
  fonttitle=\bfseries,
  boxrule=0.8pt,
  arc=3pt,
  left=8pt,right=8pt,top=6pt,bottom=8pt,
}

\theoremstyle{plain}

\theoremstyle{remark}

\usepackage{lineno}

\definecolor{darkblue}{rgb}{0, 0, 0.5}
\hypersetup{colorlinks=true, citecolor=darkblue, linkcolor=darkblue, urlcolor=darkblue}

\title{RRPO: Reference-Relative Policy Optimization with Stratified Conditional Rollouts}
\author{
  \textbf{Yuxin Xiong}$^{1}$\thanks{\;Equal contribution.} \quad
  \textbf{Xunyi Jiang}$^{1}$\footnotemark[1] \quad
  \textbf{Rohan Surana}$^{1}$\footnotemark[1] \quad
  \textbf{Xintong Li}$^{1}$ \\
  \textbf{Sheldon Yu}$^{1}$ \quad
  \textbf{Nikki Lijing Kuang}$^{1}$ \quad
  \textbf{Ryan A. Rossi}$^{2}$ \quad
  \textbf{Jingbo Shang}$^{1}$ \\
  \textbf{Tong Yu}$^{2}$ \quad
  \textbf{Julian McAuley}$^{1}$ \quad
  \textbf{Junda Wu}$^{1}$ \\
  $^{1}$University of California San Diego \quad
  $^{2}$Adobe Research \\
  \texttt{\{y7xiong, xuj003, rsurana, xil240, ziy040,} \\
  \texttt{l1kuang, jshang, jmcauley, juw069\}@ucsd.edu} \\
  \texttt{\{ryrossi, tyu\}@adobe.com}
}

\begin{document}
\maketitle
% \input{content/0_abstract}

% \input{content/1_intro}

% \input{content/5_theory}
% \input{content/6_exp}
% \input{content/7_analysis} 
% \input{content/8_conclusion}
% \bibliography{custom}
% \appendix
% \input{content/2_related}
% \input{content/9_app}
\begin{abstract}

Group Relative Policy Optimization (GRPO) has shown strong effectiveness in reinforcement learning from verifiable feedback, where sampled rollouts can be compared within a group using task-provided correctness signals. However, extending group-relative optimization beyond verifiable settings is challenging because success in many tasks is not captured by a single correctness criterion. We propose \textbf{Reference-Relative Policy Optimization (RRPO)}, which generalizes GRPO by replacing direct correctness-based advantage construction with reference-relative contrastive comparisons. RRPO first uses \emph{stratified conditional rollouts} to construct positive and negative anchor sets, and then trains a metric projection head with a set-contrastive objective to compare candidate rollouts against these anchors. The resulting alignment scores directly define contrastive advantages: during policy optimization, the projection head is frozen, and the scores are centered within each rollout group in a standard group-relative objective. We evaluate RRPO using anchor-based contrastive advantages throughout policy optimization, without relying on task ground-truth verifiers. Across verifiable reasoning, open-ended generation, and post-SFT settings, RRPO remains competitive with verifier-based optimization, improves over weakly supervised baselines, and provides additional gains after supervised fine-tuning.

% Group Relative Policy Optimization (GRPO) has shown strong effectiveness in reinforcement learning from verifiable feedback, where sampled rollouts can be compared within a group using task-provided correctness signals. However, extending group-relative optimization beyond verifiable settings is challenging because success in many tasks is contextual rather than explicitly verifiable. We propose \textbf{Reference-Relative Policy Optimization (RRPO)}, which generalizes GRPO by replacing direct correctness-based advantage construction with reference-relative contrastive comparisons. RRPO first uses \emph{stratified conditional rollouts} to construct positive and negative anchor sets, and then trains a metric projection head with a set-contrastive objective to compare candidate rollouts against these anchors. The resulting alignment scores directly define contrastive advantages: during policy optimization, the projection head is frozen, and the scores are centered within each rollout group in a standard group-relative objective. We evaluate RRPO across verifiable and open-ended benchmarks, where anchor-based contrastive advantages are used throughout policy optimization without relying on task ground-truth verifiers. Across these settings, RRPO provides a stable training signal for group-relative policy optimization beyond settings with naturally available scalar feedback.

\end{abstract}

\section{Introduction}
\label{sec:intro}

% Group Relative Policy Optimization (GRPO)~\cite{guo2025deepseek, cppo2025, yu2025dapo}, has become a practical policy-optimization tool in reinforcement learning from verifiable feedback. In such settings, sampled rollouts for the same input can be compared using task-provided correctness signals, such as exact-answer matching, symbolic checking, or execution-based tests. GRPO leverages this structure by sampling multiple rollouts per input and normalizing their scalar feedback within the group, producing relative advantages that stabilize policy updates~\cite{}.

% However, many useful generative tasks do not define success through explicit verification. Long-form explanation, open-ended question answering, dialogue summarization, and evidence-grounded generation often admit many valid outputs whose quality depends on context, coverage, style, and task intent. Even when references are available for evaluation, they do not naturally provide reliable per-rollout supervision for policy optimization: overlap-based metrics can be noisy, multiple distinct outputs may be valid, and absolute scores may not induce stable comparisons among rollouts for the same input. This creates a gap for group-relative methods: the relative update is still attractive, but the key question becomes how to construct advantages that preserve within-group comparison when correctness is not directly verifiable.
Group Relative Policy Optimization (GRPO)~\citep{shao2024deepseekmath, guo2025deepseek, lin2026cppo, wsgrpo, surana2026f, surana2026generate, mroueh2025reinforcement, yu2025dapo} 
has become a practical policy-optimization tool in reinforcement learning from verifiable feedback, comlimenting the static preference optimization~\cite{rafailov2023direct,huang2026listwise,wang2026scenealign,huang2025image,li2026importance,wucontext}. 
It is closely related to broader REINFORCE/PPO-style post-training methods for language models~\citep{schulman2017proximalpolicyoptimizationalgorithms, yu2026olivia, ahmadian2024back, massdpo, ouyang2022training, lambert2025tulu3pushingfrontiers}, but is particularly well suited to settings where multiple sampled rollouts for the same input can be compared using task-provided correctness signals. 
Examples include exact-answer matching for mathematical reasoning benchmarks such as GSM8K~\citep{cobbe2021training}, symbolic or rule-based checking in mathematical reasoning~\citep{shao2024deepseekmath, guo2025deepseek}, and execution-based tests for program-like tasks. 
GRPO leverages this structure by sampling multiple rollouts per input and normalizing their scalar feedback within the group, producing relative advantages that stabilize policy updates~\citep{shao2024deepseekmath, liu2025understanding, mroueh2025reinforcement, yu2025dapo}.

However, many useful generative tasks do not define success through explicit verification. 
Long-form explanation, open-ended question answering, dialogue summarization, and evidence-grounded generation often admit many valid outputs whose quality depends on context, coverage, style, and task intent, as reflected in datasets such as ELI5~\citep{fan2019eli5} and SAMSum~\citep{gliwa2019samsum}. 
This challenge is also central to RLHF and preference-based alignment, where supervision is typically provided through human or learned preferences rather than deterministic correctness labels~\citep{christiano2017deep, stiennon2020learning, ouyang2022training, rafailov2023direct, ethayarajh2024kto}. 
Even when references are available for evaluation, they do not naturally provide reliable per-rollout supervision for policy optimization: overlap-based metrics can be noisy, multiple distinct outputs may be valid, and absolute scores may not induce stable comparisons among rollouts for the same input~\citep{stiennon2020learning, fan2019eli5, gliwa2019samsum, tang2025beyond}.

\begin{figure*}[t]
    \centering
    \includegraphics[width=0.9\textwidth]{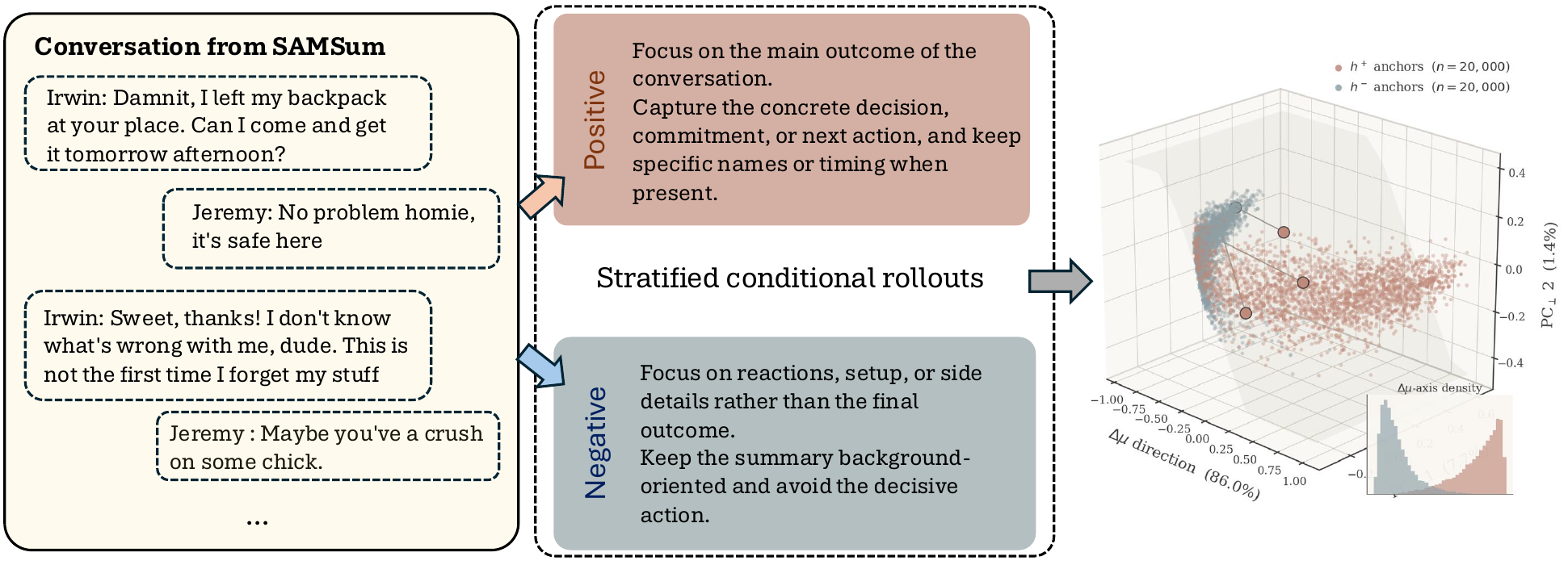}
    \caption{Stratified conditional rollouts in RRPO. For each input, weak positive and negative conditions generate instance-specific anchor sets that emphasize desirable and undesirable reference behaviors. In the SAMSum example, positive anchors focus on the main outcome of the conversation, while negative anchors emphasize peripheral or unresolved details. An offline-trained metric head then compares policy rollouts against these anchors to produce reference-relative alignment scores.}
    \label{fig:intro-overview}
    \vspace{-1em}
\end{figure*}

Recent work has therefore begun to study reinforcement learning and self-improvement beyond strictly verifiable rewards, including self-rewarding or self-training approaches~\citep{zelikman2022star, gulcehre2023reinforced, yuan2024self}, RL on unverifiable data~\citep{tang2025beyond, wang2026native}, ranked or relative reward formulations~\citep{choi2026gopo}, and self-verification or clustering-based feedback construction~\citep{zhang2026silence, jiang2025semantic}. 
This creates a gap for group-relative methods: the relative update is still attractive, but the key question becomes how to construct advantages that preserve within-group comparison when correctness is not directly verifiable.

% We propose \emph{Reference-Relative Policy Optimization} (RRPO), a weakly supervised method that computes advantages directly through contrastive learning. Rather than treating open-ended policy optimization as a problem of assigning scalar scores to individual rollouts, RRPO frames advantage estimation as reference-relative comparison. For each input, we construct positive and negative anchor sets using \emph{stratified conditional rollouts}. These anchors are generated from weak task-specific conditions that induce more desirable or less desirable behaviors, but they are not used as scalar supervision.\tong{Might it be helpful to explain, in the introduction, why reference-relative contrastive comparison is preferred over training a lightweight reward model on the same weakly conditioned anchor pairs, so that the non-triviality of our design choice is more apparent?}

We propose \emph{Reference-Relative Policy Optimization} (RRPO) as in Figure~\ref{fig:intro-overview}, a weakly supervised method that computes advantages through contrastive comparison rather than task-provided verification. Rather than assigning rollout quality in isolation, RRPO frames advantage estimation as an instance-conditioned comparison against reference behaviors. For each input, we construct positive and negative anchor sets using \emph{stratified conditional rollouts}. These anchors are generated from weak task-specific conditions that induce more desirable or less desirable behaviors, but they define only relative reference structure rather than scalar supervision.

RRPO learns to compare rollouts relative to these anchor sets. We train a metric projection head offline with a set-contrastive objective: candidate rollouts are compared against positive and negative anchors, and the projection space is optimized so that desirable rollouts align more strongly with positive anchors than with negative anchors. During policy optimization, the projection head is frozen. Each newly sampled policy rollout is assigned a reference-relative alignment score based on its aggregate contrastive alignment to the two anchor sets.

These alignment scores directly define RRPO advantages. For each input, RRPO samples a group of policy rollouts, computes their reference-relative alignment scores, and centers the scores within the group to obtain group-relative advantages. The policy is then updated with a standard clipped group-relative objective. In this way, RRPO preserves the core optimization structure of GRPO while changing the source of the advantage: advantages are induced by contrastive comparison to weakly constructed anchors rather than by task-provided verifiers.

% We evaluate RRPO in both verifiable and open-ended settings. On GSM8K, RRPO is competitive with verifier-based Dr.GRPO, showing that reference-relative contrastive advantages can remain effective even when exact verification is available. On open-ended benchmarks, where training-time verification is unavailable or unnatural, RRPO consistently improves over weakly supervised baselines. We further show that RRPO can be applied after supervised fine-tuning, yielding additional gains on top of a strong SFT initialization. These results suggest that contrastively computed reference-relative advantages provide a practical training signal for group-relative policy optimization beyond strictly verifiable tasks.\tong{Might it be helpful to qualify, in the introduction, the claim that RRPO ``consistently improves over weakly supervised baselines,'' given that Table~\ref{tab:open-results} reports missing (``--'') results for Qwen3-1.7B on ELI5?}

We evaluate RRPO in both verifiable and open-ended settings. On GSM8K, RRPO is competitive with verifier-based Dr.GRPO, showing that reference-relative contrastive advantages can remain effective even when exact verification is available. On open-ended benchmarks, where training-time verification is unavailable or unnatural, RRPO provides a more stable weak-supervision signal than pseudo-target fine-tuning and improves aggregate metrics across ELI5 and SAMSum. We further show that RRPO can be applied after supervised fine-tuning, yielding additional gains on top of a strong SFT initialization. These results suggest that contrastively computed reference-relative advantages provide a practical training signal for group-relative policy optimization beyond strictly verifiable tasks.
% \paragraph{Contributions.}
% \begin{itemize}[leftmargin=1.5em,itemsep=0.05em]
%   \item We propose RRPO, a weakly supervised extension of group-relative policy optimization that replaces verifier-based rewards with reference-relative contrastive advantages.
%   \item We introduce stratified conditional rollouts to construct instance-specific positive and negative anchor sets from weak task conditions.
%   \item We develop a set-contrastive metric objective that scores policy rollouts by their relative alignment to positive versus negative anchors, and use the resulting scores as centered group-relative advantages.
%   \item We evaluate RRPO across verifiable reasoning, open-ended generation, and post-SFT settings, showing that it is competitive with verifier-based optimization, more stable than pseudo-target fine-tuning, and complementary to supervised fine-tuning.
% \end{itemize}
\textbf{Our contributions are as follows.}
\begin{itemize}[leftmargin=1.5em,itemsep=0.2em,topsep=0.35em,parsep=0em,partopsep=0em]
  \item We propose RRPO, a weakly supervised extension of group-relative policy optimization that replaces verifier-based rewards with reference-relative contrastive advantages.
  \item We introduce stratified conditional rollouts to construct instance-specific positive and negative anchor sets from weak task conditions.
  \item We develop a set-contrastive metric objective that scores policy rollouts by their relative alignment to positive versus negative anchors, and use the resulting scores as centered group-relative advantages.
  \item We evaluate RRPO across verifiable reasoning, open-ended generation, and post-SFT settings, showing that it is competitive with verifier-based optimization, more stable than pseudo-target fine-tuning, and complementary to supervised fine-tuning.
\end{itemize}

\section{Preliminaries}
\label{sec:prelim}

% \subsection{Group-Relative Policy Optimization (GRPO)}
% \label{sec:prelim-grpo}

% Let $x\in\mathcal{X}$ denote an input instance and $\pi_\theta$ a policy over trajectories $y\in\mathcal{Y}$.
% GRPO samples a \emph{group} of $G$ rollouts $\{y_i\}_{i=1}^G$ from a snapshot policy
% $\pi_{\theta_{\mathrm{old}}}$ and constructs \emph{group-relative} advantages by centering a scalar score within the group.
% For any scalar signal $R(y;x)$, define
% \[
% A_i
% =
% R(y_i;x)
% -
% \frac{1}{G}\sum_{j=1}^{G}R(y_j;x).
% \]
% In RLVR, $R$ is usually supplied by an external verifier. In our setting, we will replace this verifier-derived score with a learned reference-relative potential.

% GRPO-style updates are typically implemented with PPO-style clipping:
% \[
% \max_{\theta}\;
% \mathbb{E}_{x}\,
% \mathbb{E}_{\{y_i\}\sim\pi_{\theta_{\mathrm{old}}}(\cdot\mid x)}
% \left[
% \frac{1}{G}\sum_{i=1}^G
% \min\!\Big(
% r_i(\theta)A_i,\,
% \mathrm{clip}(r_i(\theta),1-\epsilon,1+\epsilon)A_i
% \Big)
% \right]
% -
% \beta\,\mathrm{KL}(\pi_\theta\Vert \pi_{\mathrm{ref}}),
% \]
% where
% \[
% r_i(\theta)
% =
% \frac{\pi_\theta(y_i\mid x)}
% {\pi_{\theta_{\mathrm{old}}}(y_i\mid x)}
% \]
% and $\pi_{\mathrm{ref}}$ is a fixed reference policy.

\subsection{Group-Relative Policy Optimization (GRPO)}
\label{sec:prelim-grpo}

Let $x\in\mathcal{X}$ denote an input instance and $\pi_\theta$ a policy over trajectories $y\in\mathcal{Y}$.
GRPO samples a \emph{group} of $G$ rollouts $\{y_i\}_{i=1}^G$ from a snapshot policy
$\pi_{\theta_{\mathrm{old}}}$ and constructs \emph{group-relative} advantages by normalizing scalar scores within the group.
For any scalar signal $R(y;x)$, define the standardized group-relative advantage
\begin{equation}
\label{eq:grpo-advantage}
A_i
=
\frac{
R(y_i;x)
-
\frac{1}{G}\sum_{j=1}^{G}R(y_j;x)
}{
\operatorname{std}\left(\{R(y_j;x)\}_{j=1}^{G}\right)+\delta
}.
\end{equation}

In RLVR, $R$ is usually supplied by task-provided verification. In RRPO, this score is replaced by a learned reference-relative contrastive score.

% We use a GRPO-style clipped objective with Dr.GRPO aggregation:

% \begin{equation}
% \label{eq:drgrpo-objective}
% \begin{aligned}
% \max_{\theta}\quad
% &\mathbb{E}_{x}\,
% \mathbb{E}_{\{y_i\}_{i=1}^G\sim\pi_{\theta_{\mathrm{old}}}(\cdot\mid x)}
% \Bigg[
% \frac{1}{G L_{\max}}
% \sum_{i=1}^{G}\sum_{t=1}^{|y_i|}
% \\
% &\qquad
% \min\!\Big(
% r_{i,t}(\theta)A_i,\,
% \mathrm{clip}(r_{i,t}(\theta),1-\epsilon,1+\epsilon)A_i
% \Big)
% \Bigg]
% \\
% &\quad
% -\beta\,\mathrm{KL}(\pi_\theta\Vert \pi_{\mathrm{ref}}).
% \end{aligned}
% \end{equation}
We use a GRPO-style clipped objective with Dr.GRPO aggregation. Let
\(Y=\{y_i\}_{i=1}^G\), with \(y_i\sim\pi_{\theta_{\mathrm{old}}}(\cdot\mid x)\), and define
\(c_{i,t}(\theta)=\mathrm{clip}(r_{i,t}(\theta),1-\epsilon,1+\epsilon)\).

\begin{equation}
\label{eq:drgrpo-objective}
\begin{aligned}
\max_{\theta}\quad
&\mathbb{E}_{x,Y}
\Bigg[
\frac{1}{G L_{\max}}
\sum_{i=1}^{G}\sum_{t=1}^{|y_i|}
\\
&\qquad
\min\!\left(
r_{i,t}(\theta)A_i,\,
c_{i,t}(\theta)A_i
\right)
\Bigg]
\\
&\quad
-\beta\,\mathrm{KL}(\pi_\theta\Vert \pi_{\mathrm{ref}}).
\end{aligned}
\end{equation}

where
$
r_{i,t}(\theta)
=
\frac{\pi_\theta(y_{i,t}\mid x,y_{i,<t})}
{\pi_{\theta_{\mathrm{old}}}(y_{i,t}\mid x,y_{i,<t})},
$
\(\pi_{\mathrm{ref}}\) is a fixed reference policy, and \(L_{\max}\) is a fixed completion-length cap set before training and used as the Dr.GRPO normalizer. This differs from standard GRPO aggregation, which normalizes each rollout by its own length \(|y_i|\). Thus, our implementation uses Dr.GRPO-style fixed-length loss normalization together with group-standardized advantages; in RRPO, the key change is how the scalar signal \(R(y;x)\) is constructed.

\subsection{Metric and Contrastive Learning}
\label{sec:prelim-contrast}

Contrastive metric learning builds an embedding space in which related samples are close and unrelated samples are separated.
Given an anchor $u$, a positive sample $v^+$, and negative samples $\{v^-\}$, a canonical objective is InfoNCE:

\begin{equation}
\begin{aligned}
\mathcal{L}_{\mathrm{InfoNCE}}(u)
&=
-\log
\frac{
\exp\!\left(s(u,v^+)/\tau\right)
}{
Z(u)
}, \\
Z(u)
&=
\exp\!\left(s(u,v^+)/\tau\right)
+
\sum_{v^-}\exp\!\left(s(u,v^-)/\tau\right).
\end{aligned}
\end{equation}

where $s(\cdot,\cdot)$ is a similarity function and $\tau>0$ is a temperature.

RRPO uses the same contrastive principle, but with \emph{sets} of positive and negative anchor rollouts rather than individual samples. The resulting set-contrastive loss aggregates similarity over positive and negative anchor sets, yielding a learned metric that compares a policy rollout against instance-specific positive and negative reference behaviors.

\section{Method}
\label{sec:method}

\begin{figure*}[t]
    \centering
    \includegraphics[width=0.8\textwidth]{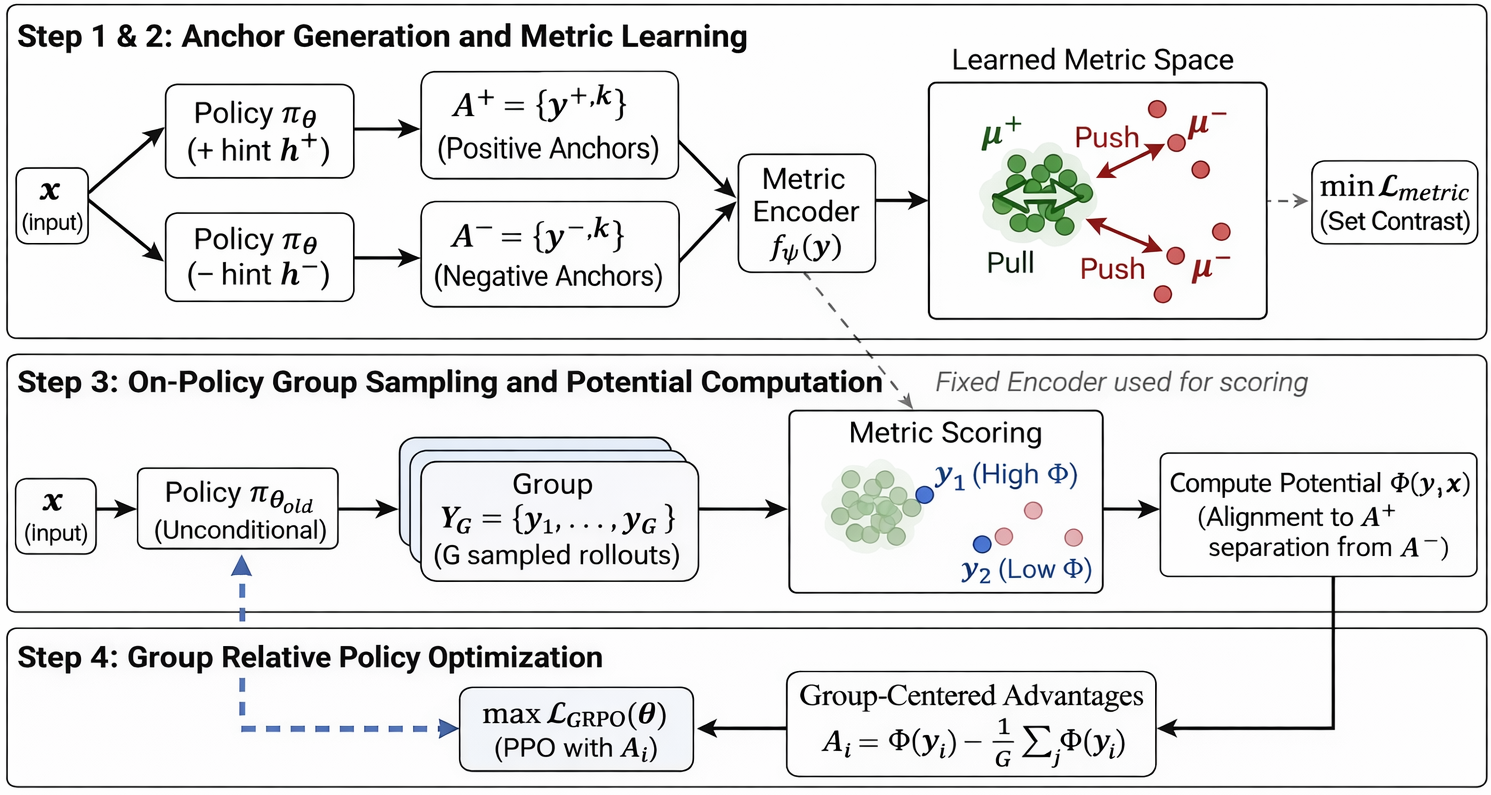}
    \caption{
    Loss function calculation during training process
    }
    \label{fig:rrpo-overview}
\end{figure*}

RRPO is a weakly supervised approach for constructing group-relative advantages when task success is not directly verifiable. The central idea is to replace direct correctness-based advantage construction with reference-relative contrastive comparison. As shown in Figure~\ref{fig:rrpo-overview}, RRPO first builds positive and negative anchor sets for each input using stratified conditional rollouts. It then trains a metric projection head offline to compare policy rollouts against these anchors. During policy optimization, the metric head is frozen, and its reference-relative alignment scores are standardized within each rollout group to produce advantages for a Dr.GRPO-style update.

Importantly, RRPO does not use task ground-truth verifiers to compute scalar rewards during policy optimization. The learned metric is only used to compare rollouts relative to weakly constructed anchor sets. This makes the supervision comparative and instance-conditional: a rollout is preferred when it is closer to positive reference behaviors than to negative ones for the same input.

% \begin{figure}[t]
%     \centering
%     \includegraphics[width=0.8\linewidth]{figures/fig2.png}
%     \caption{
%     Loss function calculation during training process
%     }
%     \label{fig:rrpo-overview}
% \end{figure}

\subsection{Stratified Conditional Rollouts}
\label{sec:stratified-rollouts}

For an input $x$, RRPO uses weak task-specific conditions to generate two strata of reference behaviors. Let $c^+(x)$ denote a positive condition that steers generation toward desirable task behavior, and let $c^-(x)$ denote a negative condition that steers generation toward less desirable behavior. We call these conditions weak because they provide only directional, task-level guidance: they are not calibrated quality scores, are not converted into scalar labels, and are not used as task verifiers.

Using a frozen anchor-generation policy $\pi_{\mathrm{anc}}$ initialized from the same base model as $\pi_{\theta_0}$, RRPO samples $K$ positive and $K$ negative anchor rollouts:
\begin{equation}
\label{eq:anchor-sets}
\begin{aligned}
\mathcal{A}^+(x) &= \{y^{+,k}\}_{k=1}^{K},
& y^{+,k} &\sim \pi_{\mathrm{anc}}(\cdot\mid x,c^+(x)), \\
\mathcal{A}^-(x) &= \{y^{-,k}\}_{k=1}^{K},
& y^{-,k} &\sim \pi_{\mathrm{anc}}(\cdot\mid x,c^-(x)).
\end{aligned}
\end{equation}
The trainable policy produces unconditional rollouts $y\sim\pi_\theta(\cdot\mid x)$, which are scored by comparison to these anchor sets. In our implementation, anchors are sampled and cached during Stage~1 metric-data construction, then reused throughout Stage~2 policy optimization. Equivalently, the anchor-generation process is fixed with respect to policy-gradient updates, so the reference structure remains stable while the policy changes.

\subsection{Offline Set-Contrastive Metric Learning}
\label{sec:metric}

RRPO learns a metric space in which candidate rollouts can be compared to the positive and negative anchor sets. Let $h_\theta(y;x)$ denote the mean-pooled hidden representation over completion tokens for the prompt-completion pair $(x,y)$, and let $g_\psi$ be an MLP projection head. The metric is trained on rollouts from the initialization policy and then used during a KL-regularized policy-optimization stage, which keeps the scored rollout distribution close to the metric-training distribution. We define the normalized trajectory embedding
$
f_\psi(y;x)
=
\frac{g_\psi(h_\theta(y;x))}
{\|g_\psi(h_\theta(y;x))\|_2}.
$
For readability, we omit $x$ when clear from context. Similarity is cosine similarity:
\[
s_\psi(y,y')=f_\psi(y;x)^\top f_\psi(y';x).
\]

The projection head is trained offline using unconditional query rollouts and stratified anchor sets generated from the same initialization policy. For each input $x$, we sample query rollouts
$
\{y_m\}_{m=1}^{G_{\mathrm{metric}}}
\sim
\pi_{\theta_0}(\cdot\mid x),
$
together with $\mathcal{A}^+(x)$ and $\mathcal{A}^-(x)$. For a query rollout $y_m$, define its aggregate alignment to each anchor set:

\begin{equation}
\label{eq:anchor-alignment}
\begin{aligned}
S_\psi^+(y_m;x)
&=
\log \sum_{k=1}^{K}
\exp\!\big(s_\psi(y_m,y^{+,k})/\tau\big),
\\
S_\psi^-(y_m;x)
&=
\log \sum_{k=1}^{K}
\exp\!\big(s_\psi(y_m,y^{-,k})/\tau\big).
\end{aligned}
\end{equation}

where $\tau>0$ is the same temperature used below in the set-contrastive loss.

The set-contrastive loss is defined over the positive-negative alignment
margin. Let \(S_m^\pm = S_\psi^\pm(y_m;x)\). Then
\begin{equation}
\label{eq:set-contrastive-loss}
\mathcal{L}_{\mathrm{sc}}(x,y_m)
=
-\log \sigma\!\left(S_m^+ - S_m^-\right).
\end{equation}

% The set-contrastive loss is
% \begin{equation}
% \label{eq:set-contrastive-loss}
% \mathcal{L}_{\mathrm{sc}}(x,y_m)
% =
% -\log
% \frac{\exp(S_\psi^+(y_m;x))}
% {\exp(S_\psi^+(y_m;x))+\exp(S_\psi^-(y_m;x))}.
% \end{equation}

% Equivalently,
% \[
% \mathcal{L}_{\mathrm{sc}}(x,y_m)
% =
% -\log \sigma\!\left(S_\psi^+(y_m;x)-S_\psi^-(y_m;x)\right).
% \]
This objective encourages query rollouts to align more strongly with positive anchors than with negative anchors. The offline metric-learning objective aggregates this loss over query rollouts and averages over training inputs:
\begin{equation}
\label{eq:metric-objective}
\min_{\psi}\quad
\mathcal{L}_{\mathrm{metric}}
=
\mathbb{E}_{x}
\left[
\sum_{m=1}^{G_{\mathrm{metric}}}
\mathcal{L}_{\mathrm{sc}}(x,y_m)
\right].
\end{equation}

After training, the projection parameters $\psi$ are frozen.

\subsection{Reference-Relative Contrastive Advantages}
\label{sec:adv}

During policy optimization, RRPO uses the frozen metric head to compute a reference-relative alignment score for each rollout:
% \begin{equation}
% \label{eq:rrpo-score}
$
R_\psi(y;x)
=
S_\psi^+(y;x)-S_\psi^-(y;x).
$
% \end{equation}
This score does not assign rollout quality in isolation; it is defined by contrastive comparison to the positive and negative anchor sets for the same input. A rollout receives a higher score when it is closer to positive anchors and farther from negative anchors in the learned metric space.

For each input $x$, the snapshot policy samples a rollout group
$
\{y_i\}_{i=1}^{G}
\sim
\pi_{\theta_{\mathrm{old}}}(\cdot\mid x).
$
RRPO constructs standardized group-relative advantages from the alignment scores:
\begin{equation}
\label{eq:rrpo-advantage}
A_i
=
\frac{
R_\psi(y_i;x)
-
\frac{1}{G}\sum_{j=1}^{G}R_\psi(y_j;x)
}{
\operatorname{std}\left(\{R_\psi(y_j;x)\}_{j=1}^{G}\right)+\delta
}.
\end{equation}
This step preserves the key group-relative principle: the update depends on how a rollout compares to other rollouts for the same input, rather than on an absolute scalar label.

\subsection{Policy Optimization}
\label{sec:objective}

RRPO updates the policy using the Dr.GRPO-style clipped objective described in Section~\ref{sec:prelim-grpo}, with the advantages defined above. The anchor-generation policy $\pi_{\mathrm{anc}}$ and metric projection head $g_\psi$ remain frozen during this stage, and only the trainable policy is updated. Thus, RRPO keeps the policy optimization backbone fixed while replacing correctness-based advantage construction with weakly supervised reference-relative contrastive advantages.

Algorithm~\ref{alg:rrpo} in Appendix~\ref{app:algorithm} summarizes the full two-stage procedure.
% Algorithm~\ref{alg:rrpo} summarizes the full two-stage procedure.

% \input{contents/4_method}

% \input{contents/5_theory}

\section{Experiments}
\label{sec:exp}

Our experiments are designed to assess whether reference-relative contrastive advantages can serve as a general optimization signal for group-relative policy learning. We consider three complementary regimes. The first regime examines tasks with explicit verification, testing whether RRPO can recover the benefits of group-relative policy improvement without using task verifiers as the scalar training signal. The second regime studies open-ended generation, where task success is contextual and not naturally reducible to exact pass/fail feedback, and examines whether reference-relative advantages remain effective under weak supervision. The third regime evaluates whether RRPO is complementary to supervised fine-tuning by applying reference-relative policy optimization after an SFT initialization.

Across experiments, RRPO uses the same two-stage pipeline: offline set-contrastive metric learning over cached positive and negative anchor representations, followed by Dr.GRPO-style policy optimization using standardized reference-relative scores as advantages.

\subsection{Experimental Setup}
\label{sec:exp-setup}

\textbf{Tasks and models.}
We choose tasks to cover both verifiable and open-ended settings. GSM8K~\citep{cobbe2021training} serves as a verifiable math reasoning benchmark with exact-answer checking. SAMSum~\citep{gliwa2019samsum} and ELI5~\citep{fan2019eli5}represent open-ended generation tasks, covering dialogue summarization and long-form explanation generation, where outputs are not naturally reducible to exact pass/fail verification during policy optimization. We evaluate Qwen2.5-1.5B-Instruct~\citep{yang2024qwen2}, Qwen3-1.7B~\citep{yang2025qwen3}, and Qwen3-4B-Instruct-2507~\citep{yang2025qwen3} to test RRPO across multiple instruction-tuned model scales.

\textbf{RRPO configuration.}
For RRPO, \(G\) denotes the number of unconditional rollouts per prompt and \(K\) denotes the number of positive and negative anchors per input. In the main reported runs, we use \(G=8\) and \(K=4\) unless otherwise specified. The metric temperature is typically \(0.1\), and the projection dimension is typically \(512\). GSM8K uses CoT-style weak conditions, SAMSum uses structural summarization conditions, and ELI5 uses explanation-style conditions. During policy optimization, RRPO does not use task verifiers or exact correctness labels as scalar rewards; instead, it constructs advantages from frozen reference-relative metric scores. Additional run-specific details are provided in Appendix~\ref{app:exp-details}.

%% include the training para and hint construction strategy

\textbf{Evaluation.}
For GSM8K, we report Pass@1/2/4/8 using the standard unbiased pass@k estimator with normalized exact-answer evaluation. For SAMSum, we report ROUGE-1/2/L, mean ROUGE F1, BLEU, and BERTScore-F1. For ELI5, we report ROUGE-1/2/L and mean ROUGE F1. Full evaluation details are provided in Appendix~\ref{app:eval-details}.

\textbf{Baselines.}
\textbf{Base} evaluates the original instruction model before policy optimization. \textbf{Dr.GRPO} is included on GSM8K as a verifier-based policy optimization baseline; it uses exact-answer correctness feedback and does not use RRPO's metric encoder or anchor cache. \textbf{\textsc{Pseudo-SFT}} is the weakly supervised baseline for open-ended tasks: it generates one greedy pseudo target per training input and fine-tunes on the resulting pseudo dataset, without anchors, metric learning, or group-relative advantages. \textbf{SFT} is a stronger supervised baseline and is reported separately in Section~\ref{sec:sft-rrpo} to distinguish fully supervised fine-tuning from the weakly supervised comparison.

% % \tong{Might it be helpful to include, in the experimental section, ablation studies over the key hyperparameters $G$ (rollout group size) and $K$ (anchor set size), to help readers understand the sensitivity of our method to these choices?} We don't have enough time to run those so we fix the settings

% \textbf{Baselines.}
% \textbf{Base} directly evaluates the original instruction model before training. \textbf{Dr.GRPO} is used on GSM8K, where exact-answer verification is available; it uses correctness-based scalar feedback and does not use the metric encoder or anchor cache. \textbf{\textsc{Pseudo-SFT}} is a weakly supervised self-training baseline for open-ended tasks: it generates one greedy pseudo target per training input and then fine-tunes on the generated pseudo dataset. \textsc{Pseudo-SFT} does not use anchors, metric learning, or group-relative advantages. \textbf{SFT} uses stronger task-specific supervision and is reported separately in Section~\ref{sec:sft-rrpo}.\tong{Might it be helpful to include, in the experimental section, additional baselines for the open-ended setting that also do not require task ground-truth verifiers, such as DPO-style preference optimization on the same weakly conditioned anchor pairs, to more thoroughly position RRPO among existing weakly supervised alternatives?}

\subsection{Verifiable Reasoning: GSM8K}
\label{sec:exp-gsm8k}

% \begin{figure}
%     \centering
%     \includegraphics[width=0.9\linewidth]{figures/fig2_eval_curves_best.pdf}
%     \caption{Evaluation metrics increase during RRPO training}
%     \label{fig:learning-curve}
% \end{figure}

\begin{figure}
    \centering
    \includegraphics[width=0.95\linewidth]{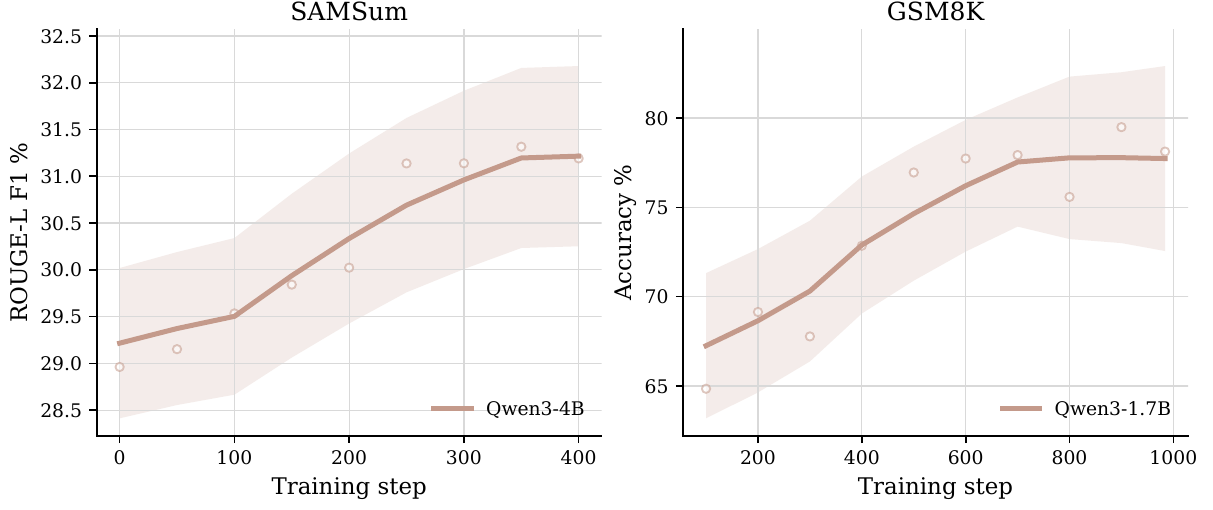}
    \caption{Learning curves during RRPO training. Evaluation performance improves over training steps on both SAMSum and GSM8K, indicating that the reference-relative training signal provides useful optimization feedback for open-ended generation and verifiable reasoning tasks.}
    \label{fig:learning-curve}
\end{figure}

\begin{table*}[t]
\centering
\small
\setlength{\tabcolsep}{5pt}
\begin{tabular*}{0.78\textwidth}{@{\extracolsep{\fill}}llcccc@{}}
\toprule
\textbf{Model} & \textbf{Method}
& \multicolumn{4}{c}{\textbf{GSM8K}} \\
\cmidrule(lr){3-6}
& & Pass@1 & Pass@2 & Pass@4 & Pass@8 \\
\midrule
Qwen2.5-1.5B & Base
& \underline{0.665} & \underline{0.781} & \underline{0.854} & \underline{0.902} \\
Qwen2.5-1.5B & Dr.GRPO
& 0.609 & 0.747 & 0.844 & 0.912 \\
Qwen2.5-1.5B & RRPO
& \textbf{0.691} & \textbf{0.807} & \textbf{0.884} & \textbf{0.934} \\
\midrule
Qwen3-1.7B & Base
& 0.746 & 0.836 & 0.891 & \underline{0.928} \\
Qwen3-1.7B & Dr.GRPO
& \underline{0.765} & \underline{0.848} & \underline{0.895} & 0.924 \\
Qwen3-1.7B & RRPO
& \textbf{0.768} & \textbf{0.849} & \textbf{0.900} & \textbf{0.938} \\
\midrule
Qwen3-4B & Base
& 0.868 & 0.913 & 0.936 & 0.944 \\
Qwen3-4B & Dr.GRPO
& \textbf{0.903} & \textbf{0.934} & \textbf{0.950} & \textbf{0.958} \\
Qwen3-4B & RRPO
& \underline{0.870} & \underline{0.918} & \underline{0.945} & \textbf{0.958} \\
\bottomrule
\end{tabular*}
\caption{GSM8K results. Dr.GRPO uses exact-answer rewards, while RRPO constructs reference-relative advantages without using ground-truth answers for policy optimization.}
\label{tab:verifiable-results}
\end{table*}

GSM8K allows a direct comparison between verifier-based and reference-relative advantage construction. Dr.GRPO uses ground-truth final answers to form sparse exact-match correctness signals, whereas RRPO does not use ground-truth final answers as policy-optimization supervision and instead derives advantages from stratified contrastive anchors.

Table~\ref{tab:verifiable-results} shows that RRPO improves over the base model across all scales. On Qwen2.5-1.5B-Instruct, RRPO raises Pass@1/8 from 0.665/0.902 to 0.691/0.934 and outperforms Dr.GRPO across all Pass@$k$ metrics. On Qwen3-1.7B, RRPO again achieves the best results, reaching 0.768 Pass@1 and 0.938 Pass@8. On Qwen3-4B-Instruct-2507, RRPO remains competitive with verifier-based Dr.GRPO and matches its Pass@8 at 0.958. These results indicate that weakly supervised reference-relative advantages can be more informative than sparse 0/1 exact-match signals for smaller models, while remaining competitive on stronger backbones.

Figure~\ref{fig:learning-curve} further shows that RRPO's GSM8K accuracy improves steadily during policy optimization, indicating that the reference-relative advantage signal provides a usable training signal rather than only improving final evaluation by chance.

\subsection{Open-Ended Generation: ELI5 and SAMSum}
\label{sec:exp-open}

\begin{table*}[t]
\centering
\small
\setlength{\tabcolsep}{2.2pt}
\begin{tabular*}{\textwidth}{@{\extracolsep{\fill}}llcccccccccc@{}}
\toprule
\textbf{Model} & \textbf{Method}
& \multicolumn{4}{c}{\textbf{ELI5}}
& \multicolumn{6}{c}{\textbf{SAMSum}} \\
\cmidrule(lr){3-6}\cmidrule(lr){7-12}
& & R-1 & R-2 & R-L & Mean-R
& R-1 & R-2 & R-L & Mean-R & BLEU & BERT \\
\midrule
Qwen2.5-1.5B & Base
& \underline{0.215} & \underline{0.036} & \underline{0.124} & \underline{0.125}
& \underline{0.408} & \textbf{0.151} & \underline{0.317} & \underline{0.292} & \underline{0.079} & \underline{0.829} \\
Qwen2.5-1.5B & \textsc{Pseudo-SFT}
& 0.210 & 0.025 & 0.102 & 0.112
& 0.295 & 0.074 & 0.214 & 0.194 & 0.030 & 0.800 \\
Qwen2.5-1.5B & RRPO
& \textbf{0.219} & \textbf{0.040} & \textbf{0.131} & \textbf{0.130}
& \textbf{0.412} & \textbf{0.151} & \textbf{0.318} & \textbf{0.294} & \textbf{0.083} & \textbf{0.830} \\
\midrule
Qwen3-1.7B & Base
& 0.198 & \textbf{0.038} & 0.116 & 0.117
& 0.382 & 0.119 & \textbf{0.293} & 0.264 & 0.065 & \textbf{0.817} \\
Qwen3-1.7B & \textsc{Pseudo-SFT}
& \textbf{0.201} & \textbf{0.038} & \textbf{0.118} & \textbf{0.119}
& 0.382 & 0.119 & 0.292 & 0.264 & \underline{0.066} & 0.816 \\
Qwen3-1.7B & RRPO
& \underline{0.200} & \textbf{0.038} & 0.116 & \underline{0.118}
& \textbf{0.383} & \textbf{0.120} & \textbf{0.293} & \textbf{0.265} & \textbf{0.067} & \textbf{0.817} \\
\midrule
Qwen3-4B & Base
& \underline{0.222} & \underline{0.034} & \underline{0.117} & \underline{0.124}
& 0.392 & 0.133 & 0.278 & 0.274 & \underline{0.070} & \underline{0.822} \\
Qwen3-4B & \textsc{Pseudo-SFT}
& 0.185 & \underline{0.034} & 0.102 & 0.107
& \underline{0.393} & \underline{0.135} & \underline{0.299} & \underline{0.276} & 0.069 & \underline{0.822} \\
Qwen3-4B & RRPO
& \textbf{0.227} & \textbf{0.039} & \textbf{0.123} & \textbf{0.129}
& \textbf{0.402} & \textbf{0.158} & \textbf{0.319} & \textbf{0.293} & \textbf{0.088} & \textbf{0.829} \\
\bottomrule
\end{tabular*}
\caption{Open-ended generation results under weak supervision. Model names are abbreviated in the table. \textsc{Pseudo-SFT} fine-tunes on model-generated pseudo targets, while RRPO constructs reference-relative advantages from positive and negative anchors.}
\label{tab:open-results}
\end{table*}

We next evaluate RRPO on ELI5~\citep{fan2019eli5} and SAMSum~\citep{gliwa2019samsum}, where training-time quality is not naturally available as an exact pass/fail verifier. This setting tests whether group-relative policy optimization can remain effective when verifier rewards are replaced by weak reference-relative comparisons.

Table~\ref{tab:open-results} compares Base, \textsc{Pseudo-SFT}, and RRPO. \textsc{Pseudo-SFT} is a natural weak-supervision baseline that fine-tunes on one model-generated pseudo target per input. RRPO uses the same weak-supervision regime differently: instead of imitating a single target, it samples multiple rollouts and constructs group-relative advantages from their contrastive alignment to positive and negative anchor sets.

RRPO is more stable than \textsc{Pseudo-SFT} across both tasks. On SAMSum, RRPO achieves the best or tied-best results for nearly all metrics and model sizes, while \textsc{Pseudo-SFT} substantially degrades the 1.5B model and gives mixed gains for larger models. The SAMSum curve in Figure~\ref{fig:learning-curve} shows the same trend during policy optimization, with ROUGE-L F1 increasing as RRPO updates proceed. On ELI5, RRPO improves Mean ROUGE over Base for all three model families, whereas \textsc{Pseudo-SFT} improves only Qwen3-1.7B and degrades the other two. These results suggest that RRPO extends group-relative optimization to unverifiable open-ended settings: weak anchors provide a useful relative advantage signal, while direct pseudo-target imitation is more brittle.

\subsection{RRPO after Supervised Fine-Tuning}
\label{sec:sft-rrpo}

\begin{table*}[t]
\centering
\small
\setlength{\tabcolsep}{2.4pt}
\begin{tabular*}{\textwidth}{@{\extracolsep{\fill}}llcccccccccc@{}}
\toprule
\textbf{Model} & \textbf{Method}
& \multicolumn{4}{c}{\textbf{ELI5}}
& \multicolumn{6}{c}{\textbf{SAMSum}} \\
\cmidrule(lr){3-6}\cmidrule(lr){7-12}
& & R-1 & R-2 & R-L & Mean-R
& R-1 & R-2 & R-L & Mean-R & BLEU & BERT \\
\midrule
Qwen2.5-1.5B & SFT
& 0.220 & 0.0280 & 0.124 & 0.124
& 0.456 & 0.190 & 0.361 & 0.336 & 0.129 & 0.853 \\
Qwen2.5-1.5B & SFT+RRPO
& 0.222 & 0.0283 & 0.126 & 0.125
& 0.456 & 0.193 & 0.363 & 0.337 & 0.131 & 0.853 \\
\midrule
Qwen3-4B & SFT
& 0.198 & 0.038 & 0.136 & 0.124
& 0.529 & 0.285 & 0.450 & 0.421 & 0.206 & 0.871 \\
Qwen3-4B & SFT+RRPO
& 0.200 & 0.038 & 0.137 & 0.125
& 0.533 & 0.287 & 0.452 & 0.424 & 0.208 & 0.872 \\
\bottomrule
\end{tabular*}
\caption{Post-SFT results on open-ended benchmarks. RRPO is applied after supervised fine-tuning using the same reference-relative advantage construction.}
\label{tab:sft-rrpo}
% \vspace{-2em}
\end{table*}

Finally, we test whether RRPO can further improve policies after conventional supervised fine-tuning. This setting is stronger than Table~\ref{tab:open-results}, since SFT already trains directly on task references. Starting from each SFT checkpoint, we apply RRPO with the same reference-relative contrastive advantage construction.

Table~\ref{tab:sft-rrpo} shows that SFT+RRPO improves or matches SFT across all reported metrics. The gains are modest but consistent: Mean ROUGE improves on both ELI5 and SAMSum for both model families, and SAMSum also improves on BLEU and BERTScore.

These results suggest that RRPO is complementary to supervised fine-tuning rather than only a weak-supervision substitute. After SFT learns from reference imitation, RRPO can still provide an additional group-relative policy-improvement signal for open-ended generation.

\textbf{Summary.}
Together, the experiments show that RRPO broadens the applicability of group-relative policy optimization beyond verifier-based training. On GSM8K, RRPO remains competitive with verifier-based Dr.GRPO without using exact-answer rewards for policy optimization. On ELI5 and SAMSum, it provides a more stable weak-supervision signal than pseudo-target fine-tuning in open-ended settings. Finally, RRPO also improves SFT-initialized policies, suggesting that reference-relative advantages can serve as a complementary policy-improvement signal beyond supervised imitation.

\section{Related Work}
\label{sec:related}

\textbf{RL from verifiable rewards and GRPO.}
Reinforcement learning from verifiable rewards (RLVR) has become a central paradigm for improving language-model reasoning in domains where task outcomes can be automatically checked. 
% DeepSeekMath introduced GRPO, a memory-efficient alternative to PPO that replaces value-function estimation with within-group reward normalization~\citep{shao2024deepseekmath}, and DeepSeek-R1 scaled this recipe to elicit emergent reasoning from correctness-based feedback~\citep{guo2025deepseek}.
Prior work introduced GRPO as a memory-efficient alternative to PPO that replaces value-function estimation with within-group reward normalization~\citep{shao2024deepseekmath}, and later work scaled this training recipe to elicit emergent reasoning from correctness-based feedback~\citep{guo2025deepseek}.
Subsequent work has refined GRPO along several axes, including stabilization at scale~\citep{yu2025dapo}, removal of length and standard-deviation biases~\citep{liu2025understanding}, and theoretical analyses that recast GRPO as a contrastive or process-reward objective~\citep{mroueh2025reinforcement,sullivan2025grpo}. 
RLOO offers a closely related leave-one-out baseline within the same group-relative family~\citep{ahmadian2024back}. 
% RRPO builds on this group-relative principle and inherits the Dr.~GRPO-style aggregation, but replaces verifier-derived rewards with reference-relative potentials induced by stratified conditional anchors.

% Reinforcement learning from verifiable rewards (RLVR) has become a central paradigm for improving language-model reasoning in domains where task outcomes can be automatically checked, such as mathematics, coding, and tool use. DeepSeekMath introduced Group Relative Policy Optimization (GRPO), a memory-efficient alternative to PPO that replaces value-function estimation with within-group reward normalization~\citep{shao2024deepseekmath}. Subsequent work has studied RLVR training dynamics, verifier design, rollout selection, and scalable reasoning improvement under exact or executable reward signals~\citep{guo2025deepseek,sheng2025hybridflow,cppo2025,yu2025dapo}. RRPO builds on the same group-relative optimization principle, but replaces verifier-derived rewards with reference-relative potentials induced by stratified conditional anchors.

\textbf{Verifier-free and relative reward signals.}
Recent work has begun extending RL-style post-training beyond domains with reliable automatic verifiers. JEPO optimizes a Jensen evidence lower bound by treating chain-of-thought as a latent variable for unverifiable or semi-verifiable data~\citep{tang2025beyond}, while Native Reasoning Training rewards latent reasoning paths that increase answer likelihood~\citep{wang2026native}. SemanticVoting replaces exact majority voting with semantic similarity voting for self-improvement on open-ended tasks~\citep{jiang2025semantic}. Other methods address unreliable scalar rewards by relying on relative or ordinal feedback: RLHF learns reward models from human preferences~\citep{ouyang2022training}, DPO directly optimizes preferred over dispreferred responses~\citep{rafailov2023direct}, and GOPO uses within-group reward rankings rather than reward magnitudes~\citep{choi2026gopo}. RRPO is also relative, but does not require preference labels, reward-model scores, or likelihood-based pseudo-rewards; it constructs positive and negative anchor sets from weak conditions and derives a scalar score from their learned metric geometry.

\textbf{Contrastive and latent-space reward construction.}
Contrastive and metric learning provide a general mechanism for learning from relative structure rather than absolute labels. Objectives such as InfoNCE and supervised contrastive learning learn representations by pulling related examples together and separating unrelated examples~\citep{oord2018representation,khosla2020supervised}. Recent verifier-free RL work also explores latent-space reward construction; for example, Latent-GRPO derives intrinsic rewards from geometric clustering properties of terminal token representations, replacing external verifiers with latent-space structure~\citep{zhang2026silence}. RRPO similarly uses representation geometry, but its reward is explicitly reference-relative: an offline metric head compares each policy rollout to positive and negative anchor sets, producing the potential used for group-centered GRPO advantages.

\section{Conclusion}

We introduced Reference-Relative Policy Optimization (RRPO), a weakly supervised extension of group-relative policy optimization for tasks without exact training-time verifiers. RRPO constructs instance-specific positive and negative anchors with stratified conditional rollouts, learns an offline set-contrastive metric over these anchors, and uses the resulting reference-relative scores as group-centered advantages for Dr.GRPO-style policy optimization. This design preserves the relative-update structure of GRPO while replacing verifier-derived rewards with comparisons to weakly constructed reference behaviors. Across verifiable reasoning, open-ended generation, and post-SFT settings, RRPO remains competitive with verifier-based optimization, provides a more stable signal than pseudo-target fine-tuning, and further improves SFT-initialized policies. These results suggest that reference-relative advantages offer a practical route for applying GRPO-style reinforcement learning to broader generative tasks beyond strictly verifiable settings.

% The paper has a section titled “Limitations” placed as follows: ACL formatting guidelines. This section can only include the discussion of limitations; strictly no new experiments, figures or analysis.

% \section{Limitations}
% \label{sec:limitations}

% RRPO relies on weak task-specific conditions to produce useful positive and negative anchor sets, so performance may depend on how well these conditions separate desirable from undesirable behaviors. Designing such conditions is less demanding than building task verifiers, but it remains dataset-dependent and may require task knowledge. The metric head is trained offline and frozen during policy optimization; although KL regularization helps limit policy drift, large shifts could make the reference-relative scores less reliable. Finally, our evaluation covers a limited set of model families and tasks and relies mainly on automatic metrics for open-ended generation. Broader benchmarks, human evaluation, repeated runs, and more detailed ablations remain important directions for future work.

\section{Limitations}
\label{sec:limitations}

RRPO relies on weak task-specific conditions to produce useful positive and negative anchor sets, so performance may depend on how well these conditions separate desirable from undesirable behaviors. Designing such conditions is less demanding than building task verifiers, but it remains dataset-dependent and may require task knowledge. The metric head is trained offline and frozen during policy optimization; although KL regularization helps limit policy drift, large shifts could make the reference-relative scores less reliable. Finally, our evaluation covers a limited set of model families and tasks and relies mainly on automatic metrics for open-ended generation. These choices limit the extent to which the results establish robustness across broader benchmarks, human preferences, random seeds, and alternative design choices.

\textbf{LLM usage disclosure:}
AI tools were used for minor writing, formatting, and visual editing assistance for generating illustrative figures

\bibliographystyle{acl_natbib}
\bibliography{main}

%%%%%%%%%%%%%%%%%%%%%%%%%%%%%%%%%%%%%%%%%%%%%%%%%%%%%%%%%%%%
\clearpage
\appendix

\section{Algorithm}
\label{app:algorithm}
Algorithm~\ref{alg:rrpo} summarizes the full two-stage procedure.

% \begin{algorithm}[t]
\begin{algorithm}[H]
\caption{Reference-Relative Policy Optimization}
\label{alg:rrpo}
\begin{algorithmic}[1]
\REQUIRE Dataset $\mathcal{D}$; metric query count $G_{\mathrm{metric}}$; rollout group size $G$; anchor size $K$; temperature $\tau$; clip parameter $\epsilon$; KL coefficient $\beta$.
\STATE Initialize trainable policy $\pi_{\theta_0}$, frozen anchor policy $\pi_{\mathrm{anc}}$ from the same base model, metric head $g_\psi$, and reference policy $\pi_{\mathrm{ref}}$.

\STATE \textbf{Stage 1: Offline set-contrastive metric learning}
% \FORALL{$x\in\mathcal{D}_{\mathrm{metric}}$}
\FORALL{$x\in\mathcal{D}$}
    \STATE Sample query rollouts $\{y_m\}_{m=1}^{G_{\mathrm{metric}}}\sim\pi_{\theta_0}(\cdot\mid x)$.
    \STATE Sample anchors $\mathcal{A}^+(x)=\{y^{+,k}\}_{k=1}^{K}\sim\pi_{\mathrm{anc}}(\cdot\mid x,c^+(x))$ and $\mathcal{A}^-(x)=\{y^{-,k}\}_{k=1}^{K}\sim\pi_{\mathrm{anc}}(\cdot\mid x,c^-(x))$.
    \STATE Cache anchor representations and update $\psi$ using $\sum_m \mathcal{L}_{\mathrm{sc}}(x,y_m)$.
\ENDFOR
\STATE Freeze metric projection head $g_\psi$.

\STATE \textbf{Stage 2: Policy optimization with frozen reference comparison}
\FOR{each update step}
  \STATE Sample minibatch $\mathcal{B}\subset\mathcal{D}$ and set $\theta_{\mathrm{old}}\leftarrow\theta$.
  \FORALL{$x\in\mathcal{B}$}
    \STATE Sample rollout group $\{y_i\}_{i=1}^{G}\sim\pi_{\theta_{\mathrm{old}}}(\cdot\mid x)$ and retrieve cached anchor representations.
    \STATE Compute $R_\psi(y_i;x)=S_\psi^+(y_i;x)-S_\psi^-(y_i;x)$ for all $i$ with frozen $g_\psi$.
    \STATE Standardize $\{R_\psi(y_i;x)\}_{i=1}^{G}$ within the group to obtain advantages $\{A_i\}_{i=1}^{G}$.
  \ENDFOR
  \STATE Update $\theta$ with the Dr.GRPO-style clipped objective using $\{A_i\}$.
\ENDFOR
\end{algorithmic}
\end{algorithm}

\section{Experiment Details}
\label{app:exp-details}

\subsection{Condition Construction}
\label{app:hints}

RRPO constructs positive and negative anchor rollouts from weak task-specific conditions. Across datasets, the positive condition \(c^+\) is designed to bias the anchor-generation policy toward more task-aligned behavior, while the negative condition \(c^-\) biases it toward less useful or less task-aligned behavior. These conditions are used only to generate anchor rollouts. During policy optimization, RRPO does not convert hints, references, or exact-answer checks into scalar verifier rewards. Table~\ref{tab:hints} summarizes the dataset-specific positive and negative conditions used to construct anchor rollouts.

\begin{table*}[t]
\centering
\small
\setlength{\tabcolsep}{4pt}
\begin{tabular}{p{0.13\linewidth}p{0.39\linewidth}p{0.42\linewidth}}
\toprule
\textbf{Dataset} & \textbf{Positive condition \(c^+\)} & \textbf{Negative condition \(c^-\)} \\
\midrule
GSM8K &
CoT-style weak condition that encourages step-by-step mathematical reasoning, such as partial or masked reference reasoning. &
Perturbed CoT-style condition, such as corrupted intermediate reasoning or misleading answer guidance. \\

ELI5 &
Explanation-style partial condition that encourages faithful, helpful long-form explanation behavior. &
Corrupted or less faithful explanation-style condition, such as perturbed reasoning or generic incorrect causal framing. \\

SAMSum &
Structure-focused summarization condition emphasizing salient dialogue content, including the main issue, decision, commitment, outcome, or next action. &
Structure-focused negative condition emphasizing less useful context, such as setup, reactions, unresolved framing, or peripheral details. \\
\bottomrule
\end{tabular}
\caption{Dataset-specific weak conditions used for anchor generation. Conditions are used only to induce positive and negative anchor rollouts.}
\label{tab:hints}
\end{table*}

\subsection{Training Details}
\label{app:training-details}

RRPO follows a two-stage offline pipeline. In Stage~1, the base policy is frozen and only the metric projection head is trained using unconditional query rollouts and positive/negative anchor rollouts. Stage~1 saves the learned projection head and cached anchor representations. In Stage~2, the metric head and anchor representations are frozen, and the policy is updated with Dr.GRPO using standardized reference-relative scores as advantages. Table~\ref{tab:rrpo-training-details} reports the main training configuration used for each dataset.

\begin{table*}[t]
\centering
\small
\setlength{\tabcolsep}{6pt}
\begin{tabular}{lccc}
\toprule
\textbf{Field} & \textbf{GSM8K} & \textbf{ELI5} & \textbf{SAMSum} \\
\midrule
Training samples & 4k & 4k & 5k \\
Rollout group size \(G\) & 8 & 8 & 8 \\
Anchors per stratum \(K\) & 4 & 4 & 4 \\
Condition mode & CoT & ELI5-CoT & Structure \\
Metric temperature \(\tau\) & 0.1 & 0.1 & 0.1 \\
Projection dimension & 512 & 512 & 512 \\
Max prompt length & 512 & 640 & 768 \\
Max completion length & 1024 & 384 & 256 \\
Policy loss & Dr.GRPO & Dr.GRPO & Dr.GRPO \\
\bottomrule
\end{tabular}
\caption{Main RRPO training configuration used in the reported experiments.}
\label{tab:rrpo-training-details}
\end{table*}

For open-ended tasks, we additionally compare against Pseudo-SFT. Pseudo-SFT first generates one greedy pseudo target per training input using the base model, then fine-tunes the same model on these generated targets with standard supervised learning. It does not use anchor sets, metric learning, group-relative advantages, or Dr.GRPO updates. Pseudo-SFT uses one epoch, learning rate \(2\times 10^{-5}\), greedy test decoding, and the same task-specific evaluation pipeline as other methods.

For the SFT baseline, we fine-tune the base model on the same task-specific training split used by RRPO with standard supervised learning, then evaluate using the same decoding and evaluation pipeline as the other methods. SFT+RRPO initializes RRPO policy optimization from the SFT checkpoint while keeping the RRPO anchor construction and metric-learning procedure unchanged.

\subsection{Evaluation Details}
\label{app:eval-details}

For GSM8K, we evaluate generated solutions using normalized exact-answer matching. The answer extractor prioritizes GSM8K-style final answers marked by \texttt{\#\#\#\#}, supports boxed answers, and normalizes extracted answers before comparison. We report the standard unbiased pass@\(k\) estimator:
\[
\mathrm{Pass@}k
=
1-\frac{\binom{n-c}{k}}{\binom{n}{k}},
\]
where \(n\) is the number of generated samples for a prompt and \(c\) is the number of correct samples. We report Pass@1, Pass@2, Pass@4, and Pass@8, averaged over test prompts.

For SAMSum, model completions are post-processed with a fixed summary extractor before evaluation. If a completion contains an explicit summary marker, the text after the marker is retained; otherwise, the first non-empty paragraph is used, with boilerplate and extra whitespace removed. We report ROUGE-1, ROUGE-2, ROUGE-L, mean ROUGE F1, BLEU, and BERTScore-F1.

For ELI5, we evaluate generated explanations against references using ROUGE-1, ROUGE-2, ROUGE-L, and mean ROUGE F1. We do not report pass@\(k\) for ELI5 because exact binary correctness is not a natural evaluation signal for long-form explanation generation.

\end{document}